\pdfoutput=1
\documentclass[conference]{IEEEtran}
\IEEEoverridecommandlockouts
\usepackage{cite}
\usepackage{amsmath,amssymb,amsfonts}
\usepackage{bm}
\usepackage{graphicx}
\usepackage[caption=false,font=footnotesize]{subfig}
\usepackage{booktabs}
\usepackage{multirow}
\usepackage{xcolor}
\usepackage{url}
\urldef{\videourl}\url{https://youtu.be/nskukDD1MFg}

\newcommand{\cbeta}{\beta_{\mathrm{c}}}

\begin{document}

\title{Sideslip-Conditioned Drift Trajectory Planning\\with Flow Matching and Control Barrier Functions}

\author{Seokjun Lee and Seung-Hyun Kong*%
\thanks{*Corresponding author (email: skong@kaist.ac.kr)}%
\thanks{Seokjun Lee and Seung-Hyun Kong are with the CCS Graduate School of Mobility, Korea Advanced Institute of Science and Technology, Daejeon, Korea, 34051 (email: \{seokjunlee, skong\}@kaist.ac.kr)}}

\maketitle

\begin{abstract}
Drifting, the deliberate control of a vehicle at large sideslip angles, is relevant to both motorsport and recovery from unexpected oversteer. Existing autonomous drifting methods typically determine drift intensity through predefined reference trajectories, optimization objectives, or learned policy rewards, rather than accepting a desired sideslip angle as an explicit input. 
We propose, to the best of our knowledge, the first generative drift trajectory planner explicitly conditioned on a requested peak sideslip angle.
A conditional flow matching model is trained on demonstrated drift laps with hindsight-derived sideslip labels and refined through reinforcement fine-tuning. A control barrier function (CBF) correction is then applied to satisfy model-based road, tire friction, and actuator constraints at the planning waypoints. In closed-loop experiments in the racing simulator Assetto Corsa, a nonlinear model predictive controller tracks the generated plans for six requested sideslip angles from $20^\circ$ to $45^\circ$, with 30 laps (approximately 75\,km) per request. The median peak sideslip angle measured across drift-corner passes follows the requested angle, with a mean absolute error of $1.2^\circ$ across the six conditions. Ablation experiments show that reinforcement fine-tuning improves conditioning accuracy, while CBF correction substantially reduces off-road and spin events.
\end{abstract}


\section{Introduction}

Drifting is a driving technique in which the driver deliberately saturates the rear tires and sustains a large sideslip angle by counter-steering. In motorsport, drifting is used to corner quickly on low-friction surfaces such as gravel and snow~\cite{tavernini2013mintime}, and it is the subject of dedicated competitions in which drivers are scored on their drift angle and line. Beyond motorsport, sideslip control is also relevant to vehicle safety. Road icing, hydroplaning, or collisions can induce sudden oversteer, requiring the driver to regain vehicle stability~\cite{voser2010drift,goh2024nmpc,lee2025oversteer}. Since ordinary drivers do not acquire the sophisticated skill required to control the sideslip angle, autonomous drifting has been actively studied to enhance vehicle safety~\cite{hindiyeh2014drifting,goh2020drifting,weber2024drifting,djeumou2023drifting,zhou2025adaptive}. An autonomous vehicle that is able to drift can take over the control from the human driver in an emergency to stabilize the slip, and can keep driving autonomously even in slip conditions~\cite{lee2025oversteer,djeumou2024referencefree}.

\begin{figure}[t]
\centering
\subfloat[Grip driving]{\includegraphics[width=0.32\columnwidth]{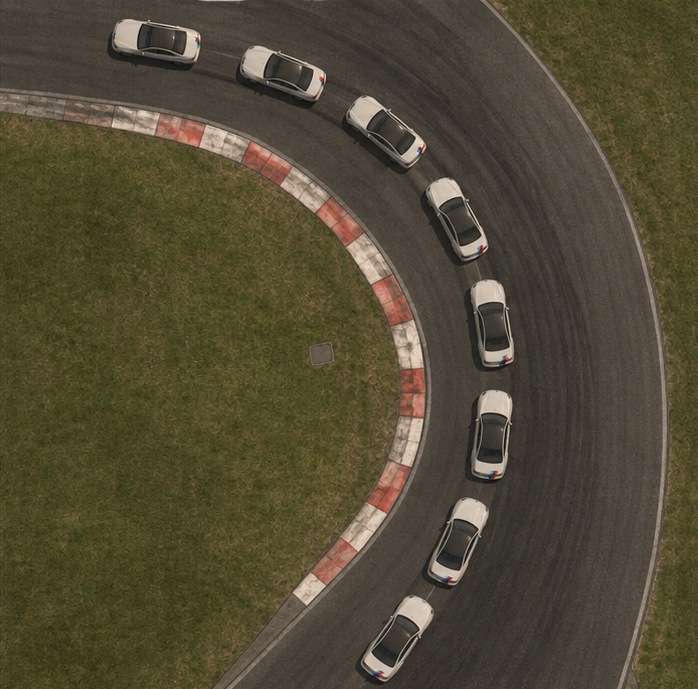}}\hfill
\subfloat[Shallow drift]{\includegraphics[width=0.32\columnwidth]{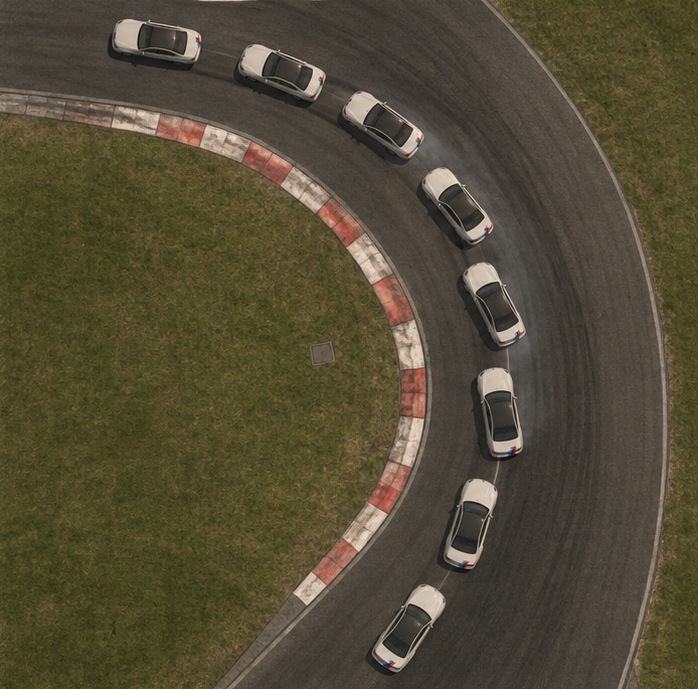}}\hfill
\subfloat[Deep drift]{\includegraphics[width=0.32\columnwidth]{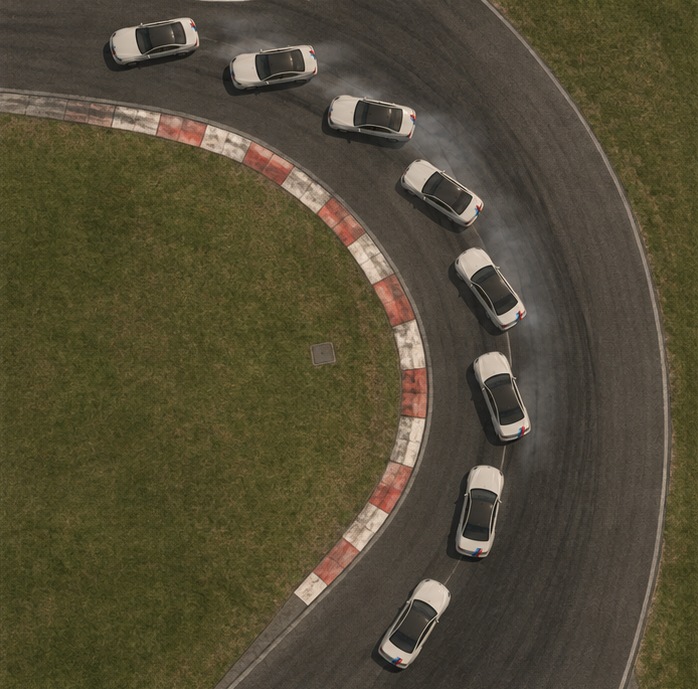}}
\caption{Illustration of grip driving (a) and shallow (b) and deep drifting (c) through the same corner.}
\label{fig:gripdrift}
\end{figure}

What distinguishes drifting from conventional driving is the ability to control the sideslip angle as an additional degree of freedom. An expert driver can take the same corner with a shallow or a deep drift, as illustrated in Fig.~\ref{fig:gripdrift}. We therefore regard the ability to control the sideslip angle at will as the core of the drifting skill.

Existing autonomous drifting methods, however, do not provide this capability, and it is difficult to add within their frameworks. Model-based methods stabilize the vehicle around a drift equilibrium or track a drift trajectory that is designed in advance or obtained from an optimal control problem~\cite{voser2010drift,hindiyeh2014drifting,goh2016drift,goh2020drifting,weber2024drifting,goh2024nmpc,djeumou2023drifting,zhou2025adaptive,tavernini2013mintime,kobayashi2024thermo}. The sideslip angle is prescribed by this reference or results from the cost function, so that changing it means designing a new reference or re-solving the problem. Learning-based methods train a drifting policy by reinforcement learning, in some cases combined with imitation learning~\cite{cutler2016drifting,cai2020drifting,domberg2022deepdrifting,djeumou2024referencefree,zhou2025psf,lee2025oversteer}. The drift behavior of such a policy is determined implicitly by its training reward, and the trained policy offers no input through which the sideslip angle can be adjusted. In both cases, the sideslip angle is fixed before deployment instead of being adjustable on request.

Recently, advances in generative models have made it possible to overcome this limitation. Diffusion and flow matching planners~\cite{janner2022diffuser,chi2023diffusionpolicy,lipman2023flow,zheng2025diffusionplanner} learn distributions of demonstrated trajectories and generate samples conditioned on inputs specified at inference time. By conditioning on the desired sideslip angle, these models can generate trajectories with different drift intensities for the same corner. However, generated trajectories are not guaranteed to satisfy road boundaries or vehicle dynamics constraints, particularly near the tire friction limit. Trajectory correction is therefore needed to improve feasibility while limiting deviations from the requested sideslip angle.

In this paper, to bridge this gap, we propose a sideslip-conditioned drift trajectory planner that combines conditional flow matching, reinforcement fine-tuning, and a control barrier function (CBF)~\cite{ames2019cbf} correction. The main contributions are as follows:
\begin{itemize}
\item We propose a sideslip-conditioned drift trajectory planner. To the best of our knowledge, it is the first generative drift planner that produces trajectories with different drift intensities for the same corner according to a requested peak sideslip angle.
\item We enforce physical and road constraints on the generated trajectories. A CBF correction accounts for road boundaries, tire friction limits, and vehicle actuator limits, including engine power and steering authority, while limiting deviations from the requested sideslip angle.
\item We validate the proposed planner in closed-loop experiments using Assetto Corsa, a high-fidelity racing simulator. Combined with a nonlinear model predictive controller (NMPC), the planner enables a test vehicle to negotiate racetrack corners at different drift intensities according to requested peak sideslip angles from $20^\circ$ to $45^\circ$.

\end{itemize}

\begin{figure*}[t]
\centering
\includegraphics[width=\textwidth]{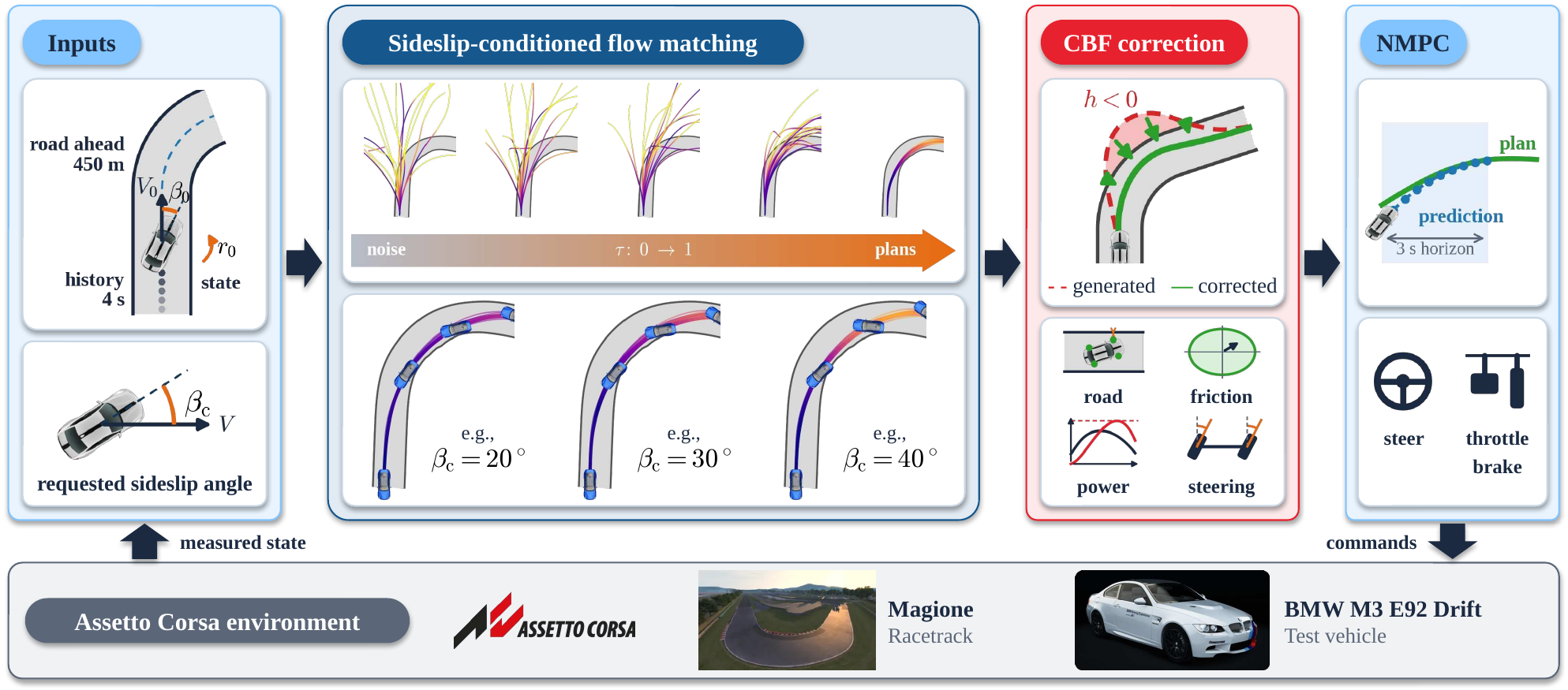}
\caption{Overview of the proposed sideslip-conditioned drift trajectory planner. Flow matching generates candidate trajectories, the CBF correction addresses the vehicle constraints, and the NMPC tracks the selected trajectory.}
\label{fig:overview}
\end{figure*}

\section{Related Work}\label{sec:related}

\subsection{Autonomous drifting}

\textbf{Model-based drift controllers} assume that the drift to be executed is given and compute the control inputs that execute it. Early studies analyzed the drift equilibria of a single-track model with nonlinear tires~\cite{voser2010drift} and stabilized them experimentally on a full-size vehicle~\cite{hindiyeh2014drifting}, and subsequent work combined equilibrium stabilization with path tracking~\cite{goh2016drift,goh2020drifting}. For transient drifts, NMPC with coupled-slip tire models and wheel-speed dynamics has been developed~\cite{weber2024drifting,goh2024nmpc}, and the model error at the friction limit has been reduced with learned tire models~\cite{djeumou2023drifting} and adaptive learning-based model predictive control~\cite{zhou2025adaptive}. In these controllers, the sideslip angle is prescribed by the given equilibrium or reference.

\textbf{Model-based drift planners} compute the drift trajectory itself by solving an optimal control problem, for example a minimum-time cornering problem~\cite{tavernini2013mintime} or the construction of a trajectory from a sequence of drift quasi-equilibria~\cite{kobayashi2024thermo}. In these approaches, the drift intensity is determined by the optimization objective and its constraints, and a drift of a different intensity requires reformulating and re-solving the problem.

\textbf{Learning-based methods} train drifting policies by reinforcement learning in simulation and on scale and full-size vehicles~\cite{cutler2016drifting,cai2020drifting,domberg2022deepdrifting}. Djeumou et al.~\cite{djeumou2024referencefree} remove the need for a reference trajectory by rewarding tire energy, Zhou et al.~\cite{zhou2025psf} learn the reference curvature of a model-based drift controller under a predictive safety filter, and Lee and Kong~\cite{lee2025oversteer} learn an end-to-end policy that recovers from sudden oversteer while avoiding obstacles. The drift behavior of such a policy is determined by its training reward, and the policy has no input through which the sideslip angle can be adjusted. Generative models have also been investigated for learning vehicle dynamics during drifting~\cite{djeumou2024onemodel}. However, the generation of drift trajectories conditioned explicitly on a requested peak sideslip angle has not been addressed.

\subsection{Generative trajectory planning}

Generative trajectory planners model the distribution of demonstrated trajectories, enabling the generation of diverse trajectories conditioned on the driving context or a specified objective. Diffusion-based methods~\cite{janner2022diffuser,chi2023diffusionpolicy} generate trajectories through iterative denoising and have been applied to autonomous driving~\cite{liao2025diffusiondrive,zheng2025diffusionplanner}. In contrast, flow matching~\cite{lipman2023flow,liu2023rectified} learns a velocity field that transports samples from a noise distribution to the data distribution through ordinary differential equation integration. Flow matching can enable efficient sampling with relatively few integration steps, particularly when trained with straight probability paths. This property is relevant to autonomous drifting, where rapid vehicle dynamics near the tire friction limit demand frequent replanning. We therefore adopt flow matching for sideslip-conditioned drift trajectory generation.

Although conditional generative models can produce trajectories for different target conditions, their training objectives do not directly optimize task-specific errors, such as deviations from desired outcomes. Recent studies have explored reinforcement fine-tuning to optimize generative policies using task-specific rewards~\cite{black2024ddpo,mcallister2025fpo,liu2025flowgrpo}.

\subsection{Constraint satisfaction in generative models}

Generative models learn to approximate the distribution of training data, but the resulting samples are not inherently guaranteed to satisfy physical or safety constraints. This limitation extends beyond autonomous driving to applications such as physical system modeling~\cite{utkarsh2025pcfm}, robot motion planning~\cite{xiao2025safediffuser}, and traffic simulation~\cite{zhong2022ctg}. Existing approaches to constraint satisfaction in generative models can be broadly categorized into three groups. Guidance-based methods steer the sampling process using cost gradients, encouraging constraint satisfaction without providing explicit guarantees~\cite{zhong2022ctg}. Projection-based methods map generated samples onto a feasible set~\cite{christopher2024projected,utkarsh2025pcfm}, while CBF-based methods incorporate barrier conditions into the sampling process to enforce safety constraints~\cite{xiao2025safediffuser,yang2026safeflowmatcher}.

Constraint satisfaction is particularly challenging in autonomous drifting, where vehicles operate near the tire friction limit. Although CBFs have been applied to filter human steering commands during drifting~\cite{dallas2025cbf}, constraint-aware generation of drift trajectories has received limited attention. Such trajectories must simultaneously satisfy road boundaries, tire friction limits, and vehicle actuator constraints, as violations can render them infeasible to track.

\section{Method}

\subsection{Problem formulation}

At each planning instant $t_0$, the planner receives the requested sideslip angle $\cbeta$, the measured vehicle state, a 4\,s history of measurements, and the road ahead. For each training demonstration, the sideslip condition is assigned in hindsight as the peak absolute sideslip angle over the 6\,s planning horizon,
\begin{equation}\label{eq:cbeta}
\cbeta = \max_{t\in(t_0,\,t_0+6\,\mathrm{s}]} |\beta(t)|,
\end{equation}
and at inference, $\cbeta$ is the requested value of this peak. The state consists of the speed $V_0$, the sideslip angle $\beta_0$, and the yaw rate $r_0$. The history contains the longitudinal speed, the longitudinal and lateral accelerations, the yaw rate, the steering angle, the throttle and brake positions, and the four wheel speeds. The road is given in the vehicle frame over the next 450\,m as its center line, left and right boundaries, and curvature. The planner outputs a plan of the vehicle position $(X, Y)$ and the yaw angle $\psi$ in the vehicle frame at $t_0$, at the waypoints $k=1,\dots,n$ over the horizon, spaced by 0.1\,s, which an NMPC tracks until the next plan becomes available (Sec.~\ref{sec:tracking}).

Each trajectory is represented by a particular solution that satisfies the initial conditions, plus a weighted sum of smooth basis trajectories extracted from demonstrations. The flow matching model generates the basis coefficients $\bm{z}$. From a plan, the course angle $\chi$ (the direction of the velocity vector), the sideslip angle $\beta=\chi-\psi$, the yaw rate $r=\dot\psi$, and the accelerations are obtained by differentiation.

To assess the dynamic feasibility of a plan, an inverse vehicle model estimates the required tire forces. The model accounts for load transfer, aerodynamic forces, and load-dependent tire friction limits identified from simulator data. For each axle $i\in\{f,r\}$, the friction utilization is
\begin{equation}
u_i=\sqrt{(F_{x,i}/\bar F_{x,i})^2+(F_{y,i}/\bar F_{y,i})^2},
\end{equation}
where $F_{x,i}$ and $F_{y,i}$ are the required axle forces and $\bar F_{x,i}$ and $\bar F_{y,i}$ are their corresponding limits. A plan satisfies the modeled friction constraint when $u_i\leq1$ for both axles. The friction utilization is used in both the reward of the reinforcement fine-tuning (RFT) and the CBF correction.

\subsection{Training}

\textbf{Supervised learning.} The generator is first trained by supervised learning (SL) as a conditional flow matching model~\cite{lipman2023flow} over the coefficients $\bm{z}$. Let $\bm{z}_1$ denote the coefficients of a demonstrated trajectory, conditioned on context $\bm{c}$ (history, road window, and state) and the sideslip condition $\cbeta$. We sample $\bm{z}_0\sim\mathcal{N}(\bm{0},\bm{I})$ as the noise input. Flow matching connects the two by the linear path
\begin{equation}
\bm{z}_\tau = (1-\tau)\,\bm{z}_0 + \tau\,\bm{z}_1,\qquad \tau\in[0,1],
\end{equation}
and trains a velocity field $v_\theta$ to regress the time derivative of this path,
\begin{equation}
\mathcal{L}_{\mathrm{SL}} = \mathbb{E}_{\tau,\,\bm{z}_0,\,(\bm{z}_1,\bm{c},\cbeta)}
\big\|v_\theta(\bm{z}_\tau,\tau,\bm{c},\cbeta) - (\bm{z}_1-\bm{z}_0)\big\|_{\bm{W}}^2,
\end{equation}
with $\tau\sim\mathcal{U}(0,1)$, where $\bm{W}$ weights the coefficients so that the loss follows the error of the trajectory. The velocity field $v_\theta$ is a transformer that encodes the history and attends to the road window (width 128, two encoder and four flow layers). A plan is generated by drawing $\bm{z}_0$ and integrating $\mathrm{d}\bm{z}/\mathrm{d}\tau = v_\theta(\bm{z},\tau,\bm{c},\cbeta)$ from $\tau=0$ to $\tau=1$ in four Euler steps.

\textbf{Reinforcement fine-tuning.} We refine the supervised generator using flow policy optimization (FPO)~\cite{mcallister2025fpo} with group-relative advantages~\cite{shao2024grpo} to improve sideslip conditioning and reduce constraint violations. For each training context, the requested sideslip angle $\cbeta$ is sampled uniformly from $0^\circ$ to $50^\circ$ if the planning horizon includes a drift corner, and set to zero otherwise. Given this request, 16 candidate trajectories are generated and evaluated without CBF correction. Each trajectory is evaluated using four reward terms:
\begin{equation}
\begin{aligned}
r_h &= -\log\Big(1+\textstyle\sum_{k,j}\max(0,-h_{k,j})^2\Big),\\
r_{\mathrm{cert}} &= \begin{cases}0 & \text{if the plan is certified},\\ -1 & \text{otherwise},\end{cases}\\
r_s &= -\log\Big(1+\tfrac{1}{n}\textstyle\sum_{k}\big[(j_{x,k}^2+j_{y,k}^2)/j_0^2+\dot r_k^2/\dot r_0^2\big]\Big),\\
r_\beta &= -\big|\max_k|\beta_k|-\cbeta\big|/10^\circ.
\end{aligned}
\end{equation}
The barrier term $r_h$ penalizes the magnitude of the violation of the barriers $h_{k,j}$ of Sec.~\ref{sec:cbf}, and $r_{\mathrm{cert}}$ penalizes a plan that does not meet the certification criterion of Sec.~\ref{sec:cbf}. The smoothness term $r_s$ penalizes the longitudinal jerk $j_x$, the lateral jerk $j_y$, and the yaw acceleration $\dot r$, which make a plan difficult to track ($j_0=5$\,m/s$^3$, $\dot r_0=1$\,rad/s$^2$). The condition term $r_\beta$ penalizes the deviation of the peak sideslip angle of the plan from the request. The reward is their weighted sum,
\begin{equation}
R = w_h r_h + w_{\mathrm{cert}} r_{\mathrm{cert}} + w_s r_s + w_\beta r_\beta,
\end{equation}
with $(w_h, w_{\mathrm{cert}}, w_s, w_\beta)=(10, 1, 0.5, 2)$. The rewards are standardized within each group to obtain advantages $A$. Following FPO, we use $\rho=\exp(\ell_{\mathrm{old}}-\ell_\theta)$ as a surrogate likelihood ratio, where $\ell$ denotes the flow matching loss evaluated at sampled $(\tau,\bm{z}_0)$. The generator minimizes
\begin{equation}
\begin{aligned}
\mathcal{L}_{\mathrm{RFT}} ={}& -\mathbb{E}\big[\min\big(\rho A,\ \mathrm{clip}(\rho,1-\epsilon,1+\epsilon)A\big)\big]\\
&+ \lambda\,\mathbb{E}\big\|v_\theta-v_{\mathrm{SL}}\big\|^2
\end{aligned}
\end{equation}
with $\epsilon=0.05$ and eight draws of $(\tau,\bm{z}_0)$ per plan. The second term keeps the velocity field close to that of the SL model. As in the adaptive penalty of~\cite{schulman2017ppo}, its weight $\lambda$ is multiplied by 1.5 after an iteration in which the term exceeds $1.5\times0.05$ and divided by 1.5 when it falls below $0.05/1.5$.

\subsection{CBF correction}\label{sec:cbf}

The RFT reward penalizes constraint violations but does not guarantee that the constraints hold. At inference, the generated plans are therefore corrected with control barrier functions (CBFs) and certified before execution. The constraints are expressed by four groups of barriers $h$, evaluated at every waypoint $k$ of the plan, and a constraint holds where its barrier is nonnegative. The barriers are normalized to comparable magnitudes.

\textbf{Road.} The four tire contact points $\bm{p}_{k,i}$ must stay between the two road boundaries,
\begin{equation}
h^{\mathrm{road}}_{k,i,s}=d_s(\bm{p}_{k,i})/w,
\end{equation}
where $d_s$ is the signed distance to the boundary $s\in\{\mathrm{left},\mathrm{right}\}$, positive on the road, and $w$ the median half-width of the road ahead.

\textbf{Friction.} The friction utilization of the front and rear axles must not exceed the admissible limit $\bar u=1$.
\begin{equation}
h^{\mathrm{fric}}_{k,i}=1-(u_{k,i}/\bar u)^2,\quad i\in\{f,r\}.
\end{equation}

\textbf{Forward motion and sideslip.} The speed must stay above $V_{\min}=1$\,m/s, and the sideslip angle is bounded by $\beta_{\max}=55^\circ$,
\begin{equation}
h^{\mathrm{fwd}}_{k}=\frac{V_k-V_{\min}}{10\,\mathrm{m/s}}
,\qquad h^{\beta}_{k}=1-(\beta_k/\beta_{\max})^2.
\end{equation}

\textbf{Actuator limits.} We impose limits on engine power and front-axle steering authority. The power limit is $P_{\max}=207$\,kW. The steering-authority constraint bounds the front-axle velocity angle $\theta_f=\mathrm{atan2}(V\sin\beta+l_fr,\,V\cos\beta)$ by $\theta_{\max}=55^\circ$, providing an approximate limit based on the vehicle's steering capability.
The corresponding barriers are
\begin{equation}
\begin{aligned}
h^{\mathrm{pow}}_{k}&=1-\max(0,F_{x,k})\,V_k/P_{\max},\\
h^{\mathrm{steer}}_{k}&=1-(\theta_{f,k}/\theta_{\max})^2.
\end{aligned}
\end{equation}

\textbf{Certification.} A plan is certified if all barriers hold after a recovery window of 1\,s and no barrier becomes worse than at the current state within this window. If the vehicle is off the road, the plan must also re-enter the road within 2\,s.

\textbf{Correction.} An uncertified plan is corrected in up to 8 steps $\bm{z}\leftarrow\bm{z}+\Delta\tau\,\Delta\bm{u}$ with $\Delta\tau=1/8$. The step $\Delta\bm{u}$ is the minimum-norm solution of the finite-time CBF condition~\cite{ames2019cbf,xiao2025safediffuser,yang2026safeflowmatcher} in the coefficient space, obtained from the quadratic program (QP)
\begin{equation}\label{eq:cbfqp}
\begin{aligned}
\min_{\Delta\bm{u},\,\bm{s}\ge\bm{0}}\ &\tfrac12\|\Delta\bm{u}\|^2+\tfrac{\kappa}{2}\|\bm{s}\|^2\\
\text{s.t.}\ &J_h\Delta\bm{u}+\bm{s}\ge-\gamma\,\mathrm{sign}(\bm{h})\odot|\bm{h}|^{1/2}.
\end{aligned}
\end{equation}
Here, $\bm{h}$ contains the active barriers ($h<0.5$), and $J_h$ is their Jacobian with respect to $\bm{z}$. We set $\gamma=5$ and $\kappa=10^4$, with slack variables $\bm{s}$ ensuring QP feasibility. The QP is solved using OSQP~\cite{stellato2020osqp}, and correction terminates upon certification.

\textbf{Selection and timing.} We generate 16 candidate trajectories, eight of which are warm-started from the previous plan. Candidates are processed in ascending order of their peak sideslip error, $|\max_k|\beta_k|-\cbeta|$. Certified candidates are then ranked by progress, sideslip accuracy, and continuity with the executed plan. The correction time budget is $\min(0.5\,\mathrm{s},\,10\,\mathrm{m}/V_0)$, with the deadline checked before each correction step.

\begin{figure*}[t]
\centering
\includegraphics[width=\textwidth]{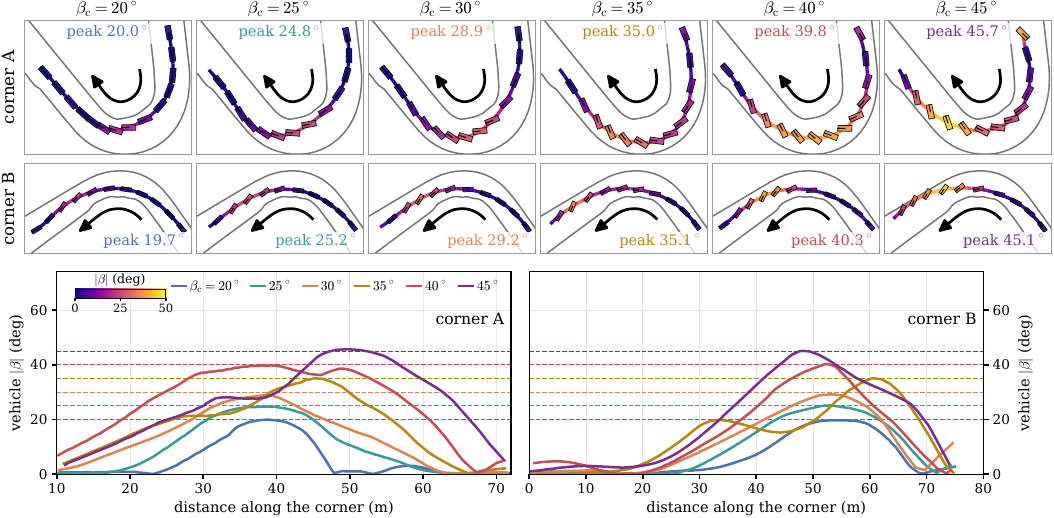}
\caption{Closed-loop drift trajectories and sideslip profiles at two corners of Magione for six requested peak sideslip angles. Vehicle poses are colored by $|\beta|$, and dashed lines indicate the requested angles.}
\label{fig:corner}
\end{figure*}

\subsection{NMPC tracking}\label{sec:tracking}

The plan is tracked by an NMPC with the single-track vehicle model of Weber and Gerdes~\cite{weber2024drifting}. Its state $\bm{x}=(X,Y,\psi,V,\beta,r,\omega_f,\omega_r,\Delta F_z,\delta,\tau)$ comprises the pose, the speed, the sideslip angle, the yaw rate, the front and rear wheel speeds, the longitudinal load transfer, the steering angle, and the wheel torque. Its input is $\bm{u}=(\dot\delta,\dot\tau)$. With $\dot X=V\cos(\psi+\beta)$, $\dot Y=V\sin(\psi+\beta)$, and $\dot\psi=r$, the body follows
\begin{align}
m\dot V ={}& F_{xf}\cos(\delta-\beta)-F_{yf}\sin(\delta-\beta)\nonumber\\
&+F_{xr}\cos\beta+F_{yr}\sin\beta-F_{\mathrm{res}},\nonumber\\
mV(\dot\beta+r) ={}& F_{xf}\sin(\delta-\beta)+F_{yf}\cos(\delta-\beta)\\
&-F_{xr}\sin\beta+F_{yr}\cos\beta,\nonumber\\
I_z\dot r ={}& l_f\,(F_{xf}\sin\delta+F_{yf}\cos\delta)-l_rF_{yr},\nonumber
\end{align}
where $m$ and $I_z$ are the mass and the yaw inertia, $l_f$ and $l_r$ the distances of the axles from the center of gravity, and $F_{\mathrm{res}}$ the aerodynamic drag and rolling resistance. The axle forces $F_{xi}$ and $F_{yi}$, $i\in\{f,r\}$, follow from a brush tire model with combined slip. Its inputs are the slip ratio, which is determined by the wheel speed $\omega_i$, the slip angles
\begin{equation}
\begin{aligned}
\alpha_f &= \mathrm{atan2}(V\sin\beta+l_fr,\;V\cos\beta)-\delta,\\
\alpha_r &= \mathrm{atan2}(V\sin\beta-l_rr,\;V\cos\beta),
\end{aligned}
\end{equation}
and the axle loads, which include the load transfer $\Delta F_z$. The tire parameters are identified from the drift laps. The NMPC uses a 3\,s horizon with 0.1\,s intervals and minimizes tracking errors in position, heading, speed, sideslip angle, and yaw rate, along with penalties on the steering and torque rates. The corresponding tracking weights are $(1, 0.1, 0.5, 1, 60, 15)$ for lateral position, longitudinal position, heading, speed, sideslip angle, and yaw rate, respectively. The optimization is solved using acados~\cite{verschueren2022acados}.

\section{Experiments}

\subsection{Setup}

\begin{figure*}[t]
\centering
\includegraphics[width=\textwidth]{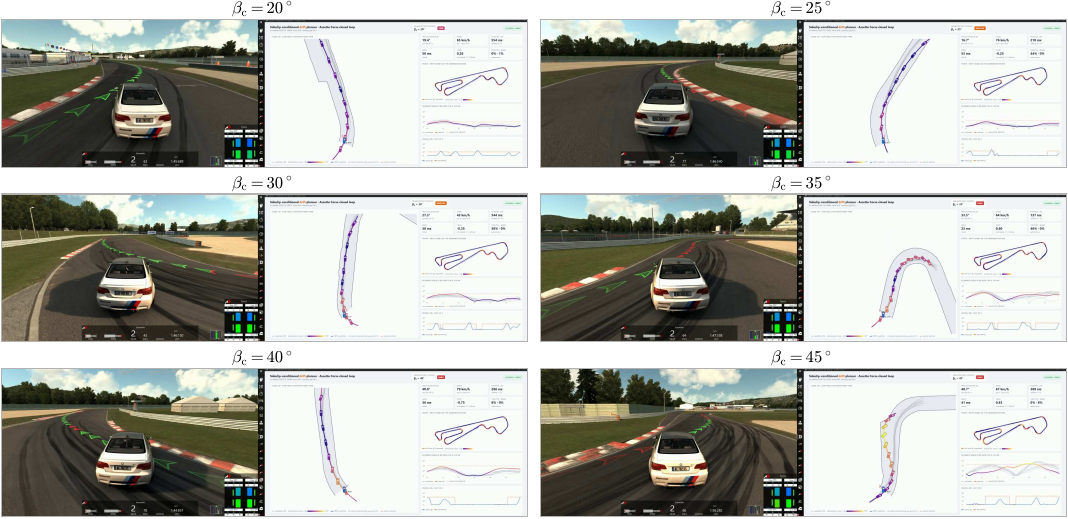}
\caption{Closed-loop drifting in Assetto Corsa for six requested peak sideslip angles. Each panel shows the vehicle chase camera and the selected trajectory. Demonstration videos can be found at \videourl.}

\label{fig:stills}
\end{figure*}

\textbf{Simulator and vehicle.} The experiments use Assetto Corsa with the BMW M3 E92 drift car on street tires, which is controlled through a virtual joystick~\cite{remonda2024acgym} at 20\,Hz. Fig.~\ref{fig:stills} shows the vehicle in the simulator for the six requests. The planner runs asynchronously on a GPU and the NMPC on the CPU, and planning takes $105$--$126$\,ms in the median. The test track is Magione, with a lap length of 2.5\,km.

\textbf{Data.} The demonstrations consist of 150,523 frames of drift laps on the test circuit, that is, about 120\,min and 73 laps. Off-track events and spins are removed ($\pm6$\,s around each). This gives 134,627 windows (4\,s history, 6\,s horizon).

\textbf{Metrics.} The condition is evaluated per corner, since $|\beta|\approx0$ on the straights is correct for any request. For each drift-corner pass, we measure the vehicle's peak absolute sideslip angle. We report the median and its 95\% confidence interval, estimated from 5,000 bootstrap resamples of corner passes. Off-road events occur when all four tires leave the road, while spins are defined by $|\beta|>90^\circ$ at speeds above 5\,m/s. Each request is evaluated over 30 laps (about 75\,km).

\subsection{Drift intensity in closed loop}\label{sec:closedloop}

\begin{table}[t]
\centering
\caption{Closed-loop performance for six requested peak sideslip angles on the Magione circuit (30 laps per request).}
\label{tab:closed}
\setlength{\tabcolsep}{2.5pt}
\begin{tabular*}{\columnwidth}{@{\extracolsep{\fill}}cccccccc@{}}
\toprule
$\cbeta$ & distance & \multicolumn{2}{c}{off-road} & \multicolumn{2}{c}{spin} & \multicolumn{2}{c}{peak $|\beta|$ ($^\circ$)} \\
\cmidrule{3-4}\cmidrule{5-6}\cmidrule{7-8}
($^\circ$) & (km) & count & /10\,km & count & /10\,km & median & 95\% CI \\
\midrule
20 & 74.3 & 0 & 0 & 0 & 0 & 16.4 & [14.6, 18.1] \\
25 & 74.4 & 1 & 0.13 & 0 & 0 & 23.0 & [19.8, 24.7] \\
30 & 74.6 & 3 & 0.40 & 0 & 0 & 28.9 & [27.1, 31.5] \\
35 & 74.7 & 2 & 0.27 & 0 & 0 & 35.1 & [32.9, 36.9] \\
40 & 75.0 & 14 & 1.87 & 0 & 0 & 39.8 & [37.3, 41.2] \\
45 & 75.1 & 21 & 2.80 & 5 & 0.67 & 44.6 & [42.7, 46.5] \\
\bottomrule
\end{tabular*}
\end{table}

Table~\ref{tab:closed} summarizes the closed-loop results for six requested sideslip angles. The median peak sideslip angle follows the request with a mean absolute error of $1.2^\circ$, and its 95\% confidence interval contains the request from $30^\circ$ to $45^\circ$. Fig.~\ref{fig:corner} further illustrates this relationship at two representative corners. Requests up to $35^\circ$ are completed without spins and with at most three off-road events over approximately 75\,km per request. However, higher sideslip requests result in more frequent road departures and occasional spins.

\begin{table}[t]
\centering
\caption{Comparison of median peak sideslip angles across planning, reference execution, and vehicle tracking.}
\label{tab:gaps}
\begin{tabular}{cccc}
\toprule
$\cbeta$ ($^\circ$) & plan ($^\circ$) & executed reference ($^\circ$) & vehicle ($^\circ$) \\
\midrule
20 & 19.7 & 18.1 & 16.4 \\
25 & 24.8 & 24.1 & 23.0 \\
30 & 29.8 & 29.7 & 28.9 \\
35 & 34.9 & 34.3 & 35.1 \\
40 & 39.9 & 38.1 & 39.8 \\
45 & 44.8 & 43.2 & 44.6 \\
\bottomrule
\end{tabular}
\end{table}

\textbf{Error analysis.} To locate the source of the remaining deviation, Table~\ref{tab:gaps} compares the median peak sideslip angles of the selected plans, the references executed by the NMPC, and the vehicle. The planned peaks are within $0.3^\circ$ of the requests, so the deviation of the vehicle does not originate in the generator. It arises in the execution of the plans. A plan is replaced by the next one after about 0.1\,s, so that only its beginning is executed, and since consecutive plans do not place the peak at the same instant, the peak of the executed reference is up to $1.8^\circ$ lower than that of the plans. The vehicle deviates from the executed reference by up to $1.7^\circ$, owing to the tracking error of the NMPC and to effects that its vehicle model does not capture, such as changes of the tire temperature and pressure and of the vehicle mass with the remaining fuel.

\subsection{Ablation study}\label{sec:ablation}

\begin{table}[t]
\centering
\caption{Closed-loop ablation study of reinforcement fine-tuning (RFT) and CBF correction.}
\label{tab:ablation}
\setlength{\tabcolsep}{2.5pt}
\begin{tabular*}{\columnwidth}{@{\extracolsep{\fill}}lccccccc@{}}
\toprule
variant & $\cbeta$ & distance & \multicolumn{2}{c}{off-road} & \multicolumn{2}{c}{spin} & peak $|\beta|$ \\
\cmidrule{4-5}\cmidrule{6-7}
 & ($^\circ$) & (km) & count & /10\,km & count & /10\,km & median ($^\circ$) \\
\midrule
RFT & 20 & 73.8 & 6 & 0.81 & 0 & 0 & 15.6 \\
 & 25 & 73.8 & 7 & 0.95 & 0 & 0 & 23.5 \\
 & 30 & 74.4 & 22 & 2.96 & 0 & 0 & 33.5 \\
 & 35 & 75.0 & 57 & 7.60 & 10 & 1.33 & 44.8 \\
 & 40 & 75.0 & 61 & 8.13 & 15 & 2.00 & 48.1 \\
 & 45 & 77.7 & 76 & 9.78 & 17 & 2.19 & 51.0 \\
\midrule
SL + CBF & 20 & 74.4 & 3 & 0.40 & 0 & 0 & 13.3 \\
 & 25 & 74.4 & 3 & 0.40 & 0 & 0 & 22.2 \\
 & 30 & 74.7 & 1 & 0.13 & 4 & 0.54 & 28.0 \\
 & 35 & 75.3 & 24 & 3.19 & 6 & 0.80 & 36.2 \\
 & 40 & 75.0 & 4 & 0.53 & 5 & 0.67 & 42.2 \\
 & 45 & 75.3 & 21 & 2.79 & 9 & 1.20 & 45.6 \\
\midrule
SL & 20 & 73.8 & 11 & 1.49 & 0 & 0 & 18.2 \\
 & 25 & 74.7 & 21 & 2.81 & 7 & 0.94 & 29.3 \\
 & 30 & 75.3 & 57 & 7.57 & 15 & 1.99 & 40.5 \\
 & 35 & 75.6 & 51 & 6.75 & 13 & 1.72 & 47.0 \\
 & 40 & 75.6 & 83 & 10.98 & 24 & 3.17 & 48.7 \\
 & 45 & 75.9 & 81 & 10.67 & 37 & 4.87 & 52.6 \\
\bottomrule
\end{tabular*}
\end{table}

To assess the contributions of RFT and CBF correction, we evaluate three ablated variants: RFT alone, supervised learning with CBF correction (SL + CBF), and supervised learning alone (SL). Each variant is evaluated over 30 laps per requested sideslip angle (Table~\ref{tab:ablation}), while the full model (RFT + CBF) is reported in Table~\ref{tab:closed}.

\textbf{Effect of CBF correction.} CBF correction substantially reduces off-road and spin events. For the RFT model, off-road events decrease from 5.09 to 0.91 per 10\,km, while spin events decrease from 0.93 to 0.11. Similar improvements are observed for the SL model, with off-road events decreasing from 6.74 to 1.25 and spin events from 2.13 to 0.53 per 10\,km. Without CBF correction, the median peak sideslip angles also exceed the requests by up to $9.8^\circ$ with RFT and $12.0^\circ$ with SL.

\textbf{Effect of RFT.} RFT improves the accuracy of sideslip conditioning. With CBF correction, the mean absolute deviation from the requested peak sideslip angle decreases from $2.6^\circ$ for SL to $1.2^\circ$ for RFT. RFT also reduces off-road and spin events compared with SL.

\section{Conclusion}

This paper presented, to the best of our knowledge, the first generative drift trajectory planner that explicitly conditions trajectory generation on a requested peak sideslip angle. By combining conditional flow matching, reinforcement fine-tuning, and CBF correction, the proposed planner enables controllable drift intensity while accounting for vehicle constraints. Closed-loop experiments in the high-fidelity racing simulator Assetto Corsa demonstrated sideslip conditioning from $20^\circ$ to $45^\circ$, with a mean absolute error of $1.2^\circ$ between the requested and median peak sideslip angles. Beyond controlled drifting, this capability provides a foundation for planning sideslip maneuvers in more complex driving tasks, such as obstacle avoidance under low-traction conditions. Future work will investigate these scenarios and extend the approach to real-world vehicle experiments.

\bibliographystyle{IEEEtran}
\bibliography{refs}

\begin{IEEEbiography}
[{\includegraphics[width=1in,height=1.25in,clip,keepaspectratio]{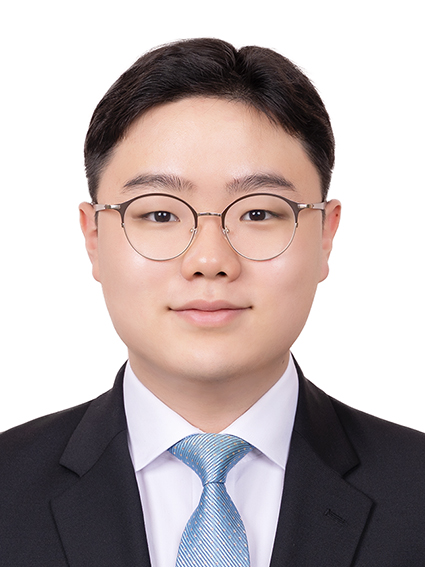}}]{Seokjun Lee} received a B.S. degree from the Department of Mechanical Engineering, Ulsan National Institute of Science and Technology (UNIST), Ulsan, South Korea, in 2023. He is currently enrolled in the Integrated M.S. and Ph.D. program at the Graduate School of Mobility, Korea Advanced Institute of Science and Technology (KAIST), Daejeon, South Korea. His research interests include end-to-end autonomous driving, autonomous driving on the handling limits, reinforcement learning and vehicle dynamics.
\end{IEEEbiography}

\begin{IEEEbiography}
[{\includegraphics[width=1in,height=1.25in,clip,keepaspectratio]{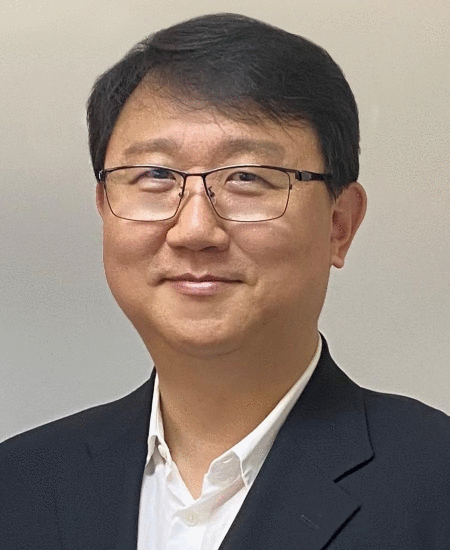}}]{Seung-Hyun Kong}(M'06--SM'16) is an Associate Professor in the CCS Graduate School of Mobility, Korea Advanced Institute of Science and Technology (KAIST), where he has been a faculty member since 2010. He received a B.S. degree in Electronics Engineering from Sogang University, Seoul, Korea, in 1992, an M.S. degree in Electrical and Computer Engineering from Polytechnic University (merged to NYU), New York, in 1994, and a Ph.D. degree in Aeronautics and Astronautics from Stanford University, Palo Alto, in 2005. From 1997 to 2004 and from 2006 to 2010, he was with companies, including Samsung Electronics (Telecommunication Research Center) and Nexpilot, both in Korea, Polaris Wireless in San Jose, and Qualcomm (Corporate R\&D) in San Diego, for advanced R\&D in mobile communication systems, wireless positioning, and assisted GNSS. His current research interests include deep neural networks for perception and localization using 4D Radar and sensor fusion, Reinforcement and Imitation learning algorithms for End-to-End autonomous driving, and neural network compression for embedded systems. He has authored more than 100 papers in peer-reviewed journals and conference proceedings and 14 patents, and his research group won the President's Award in the 2018 International Student Autonomous Driving Competition from the Korean government. He served as a Program co-Chair of IEEE ITSC 2019, New Zealand, and the General Chair of IEEE IV 2024, Korea, and he has served as an Associate Editor of IEEE T-ITS since 2017.
\end{IEEEbiography}

\vfill

\end{document}